\documentclass[10pt,twocolumn,letterpaper]{article}

\usepackage[pagenumbers]{wacv} 

\usepackage{graphicx}
\usepackage{amsmath}
\usepackage{amssymb}
\usepackage{booktabs}
\usepackage{algorithm}
\usepackage{algorithmic}
\usepackage{dsfont}

\usepackage[pagebackref,breaklinks,colorlinks]{hyperref}

\usepackage[capitalize]{cleveref}
\crefname{section}{Sec.}{Secs.}
\Crefname{section}{Section}{Sections}
\Crefname{table}{Table}{Tables}
\crefname{table}{Tab.}{Tabs.}

\usepackage{caption}
\begin{document}



\title{Vision-Language Pseudo-Labels for Single-Positive Multi-Label Learning}
\author{Xin Xing$^{1}$ \quad Zhexiao Xiong$^{2}$ \quad Abby Stylianou $^{3}$ \quad Srikumar Sastry$^{2}$ \\ \quad Liyu Gong$^4$ \quad Nathan Jacobs$^{2}$ \vspace{0.3em} \\
{\normalsize $^1$ University of Kentucky} \quad
{\normalsize $^2$ Washington University in St. Louis} \quad 
{\normalsize $^3$ Saint Louis University}
{\normalsize $^4$ Oracle Inc}
}

\maketitle

\begin{abstract}
This paper presents a novel approach to Single-Positive Multi-label Learning. In general multi-label learning, a model learns to predict multiple labels or categories for a single input image. This is in contrast with standard multi-class image classification, where the task is predicting a single label from many possible labels for an image. Single-Positive Multi-label Learning (SPML) specifically considers learning to predict multiple labels when there is only a single annotation per image in the training data. Multi-label learning is in many ways a more realistic task than single-label learning as real-world data often involves instances belonging to multiple categories simultaneously; however, most common computer vision datasets predominantly contain single labels due to the inherent complexity and cost of collecting multiple high quality annotations for each instance. We propose a novel approach called Vision-Language Pseudo-Labeling (VLPL), which uses a vision-language model to suggest strong positive and negative pseudo-labels, and outperform the current SOTA methods by 5.5\% on Pascal VOC, 18.4\% on MS-COCO, 15.2\% on NUS-WIDE, and 8.4\% on CUB-Birds. Our code and data are available at \url{https://github.com/mvrl/VLPL}.

\end{abstract}

\section{Introduction}
\label{sec:intro}
\begin{figure}[t!]
\centering
\includegraphics[width=\linewidth]{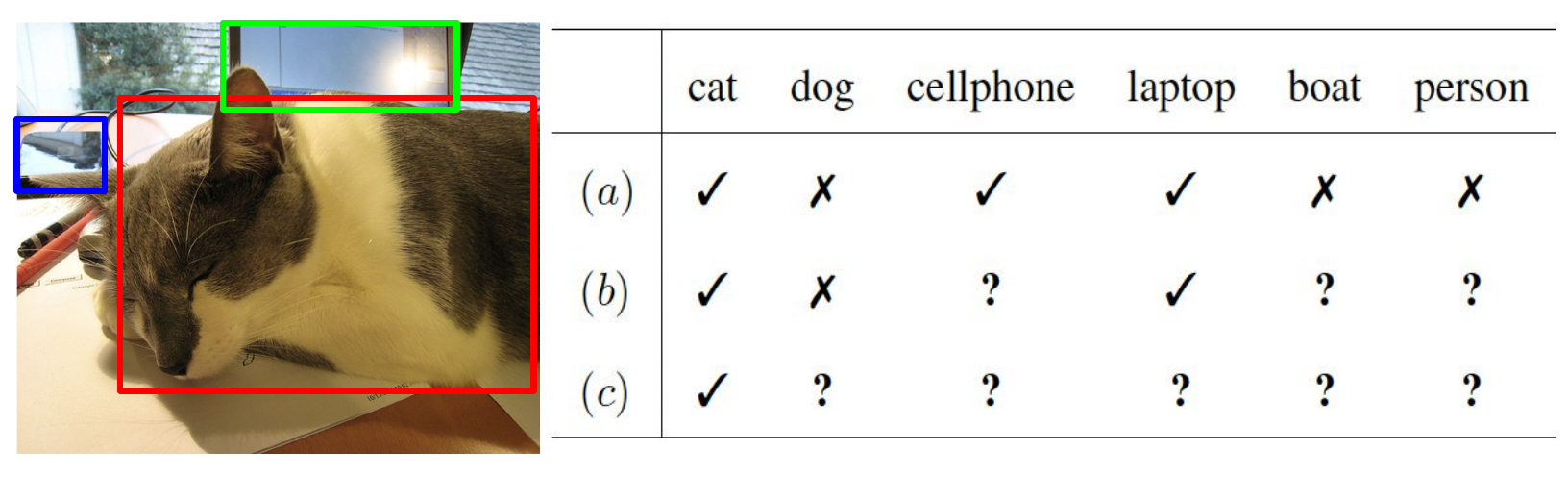}
\caption{This figure shows different levels of available annotation for multi-label learning tasks: (a) full annotation (we know all ground truth positive and negative labels), (b) partial annotation (we know partial ground truth labels, and the rest labels are unknown), and (c) single positive annotation (we only know one positive ground truth label, and the rest labels are all unknown).   }
\label{front_page}
\end{figure}

Most image classification approaches focus on performing multi-class classification: given an input image, predict which of many possible labels is the most appropriate. In Figure~\ref{front_page}, a standard image classification model would likely predict the label `cat.' Most images, however, have more than just one appropriate class. For example, Figure~\ref{front_page} shows a cat, a cell phone, and a laptop -- all of which would be appropriate labels for the image. Predicting multiple labels for an input image falls in the domain of multi-label learning. One of the largest challenges for multi-label learning is that most common computer vision datasets only provide a single annotation, even though most images contain multiple objects or classes. In~\cite{yun2021re}, the authors found that the ImageNet dataset contains 1.22 classes per image on average, even though the dataset only includes a single label per image. Collecting all possible labels for an image is time consuming, costly, and error-prone, especially when an image has a large number of classes and some of them may only be visible in a very small part of the image. This problem domain -- where the training data contains only a single label, but the task is predicting multiple labels -- is called Single-Positive Multi-Label Learning (SPML).

There are a variety of different works that focus on the SPML task~\cite{cole2021multi,taherkhani2021self,iscen2019label}. These works mainly concentrate on pseudo-labeling approaches and novel loss definitions that use these labels. Pseudo-labeling uses different types of weak supervision to identify potential positive labels for an image. These labels may come from pre-trained multi-class classification backbones, or from label-to-label association, which focuses on leveraging known or inferred relationships and dependencies between different labels. Novel losses explore how to utilize these pseudo-labels in multi-label training.

In this paper, we introduce a novel approach called Vision-Language Pseudo-Label (VLPL) for SPML. Prior pseudo-labeling work has largely focused on setting a score threshold for extracted features~\cite{durand2019learning}, or incorporating uncertainty in the pre-trained features in the pseudo-labeling~\cite{rizve2021defense}. More recently, researchers have considered using Vision-Language Models like CLIP in the pseudo-labeling process. In DualCoOp(\cite{sun2022dualcoop}), the authors use a fixed CLIP model and learn positive and negative prompt contexts per image as pseudo-labels, which are then fed into an asymmetric loss~\cite{ridnik2021asymmetric} for limited-annotation (but not single positive) multi-label classification. In~\cite{ding2023exploring}, the authors focus on incorporating label-to-label correspondence priors using a structured prior derived from a CLIP model and a Semantic Correspondence Prompt network that keys on label-to-label correspondences.

Our VLPL approach is most similar to DualCoOp. We, however, show that only using positive pseudo-labels extracted based on CLIP image-text similarity, and using an Entropy Maximization loss, can achieve SOTA performance in the SPML setting, on Pascal VOC, MS-COCO, and CUB-Birds with a significantly simpler approach. 

We demonstrate the superior performance of the VLPL model compared to baseline models, evaluate the influence of hyperparameters on the VLPL model's performance, and explore the proposed model's performance under different scenarios. The contributions of our study are as follows: 

\begin{itemize}
\setlength\itemsep{0em}
\item Inspired by the VLM application, we proposed a novel model called vision-language pseudo-label (VLPL) for SPML, which aims to produce an accurate and robust pseudo-label to boost the model performance. 

\item We conducted experiments on four benchmark multi-label datasets, i.e., PASCAL VOC~\cite{pascal-voc-2012}, MS-COCO~\cite{lin2014microsoft}, NUS-WIDE~\cite{chua2009nus}, and CUB~\cite{wah2011caltech}. Our initial experiments achieve superior performance over the baseline models, proving the effectiveness of the proposed method. Besides, by further exploration of the backbone, our method achieves new state-of-the-art results, pushing the performance boundary improvement to $5.5\%$, $18.4\%$, $15.2\%$, and $8.4\%$ over the four benchmarks, respectively.    

\item To further investigate the impact on the VLPL model's performance, we conduct more experiments to systemically evaluate our model. We examine how varying hyperparameters affected the effectiveness of the model, discuss the positive-negative imbalance in our study, and visualize the final prediction probabilities. For more details, please refer to our experiment section. 
\end{itemize}

\section{Related Work}
\subsection{Loss-function Focused Methods}
For SPML tasks, much of the work focuses on developing novel loss functions to train models. Assume Negative (AN) Loss~\cite{cole2021multi} is a simple method that assumes all the unknown labels are negative, inevitably introducing some number of false negatives in the implementation. Though the performance of AN is unsatisfactory, AN is still a widely used baseline for comparison. Entropy-Maximization (EM) loss~\cite{zhou2022acknowledging} leverages the idea of acknowledging unknown labels and aims to maximize the entropy of predicted probabilities for unannotated labels. The Weak Assume Negative (WAN) loss~\cite{mac2019presence, cole2021multi} is an updated AN method, wherein the negative labels are weighted by a `weak' coefficient to reduce the impact of false negatives. Regularized Online Label Estimation (ROLE)~\cite{cole2021multi} mirrors the expectation-maximization algorithm in jointly training the image classifier and concurrently estimating potential labels online.  Large Loss (LL)~\cite{kim2022large} proposes to overcome the memorization effect, which the model first learns the representation of clean labels, and then starts memorizing noisy labels. Our study does not primarily focus on the loss function -- we use the EM loss, which is suitable for our model as our label prediction also includes a number of unknown labels.

\subsection{Pseudo Label Focused Methods}
Pseudo-labeling is another popular method to overcome the problem of limited annotations per image. Given the imbalance between positive and negative labels (where the number of negative labels significantly outnumbers positive labels), an intuitive approach is to sample a portion of the unknown labels and assign them as negative ``pseudo-labels''~\cite{pseudo2013simple}. Clustering methods like~\cite{taherkhani2021self} leverage a distance metric to facilitate weakly-supervised or self-supervised learning for pseudo-labeling. Asymmetric Pseudo-Labeling (APL)~\cite{zhou2022acknowledging} assigns positive and negative pseudo-labels with asymmetric tolerance. This approach is often employed in conjunction with the previously mentioned EM loss. Unlike the prior pseudo-labeling methods that lack robust reference, our strategy involves leveraging an aligned vision-language embedding space to predict positive and negative labels. Our method demonstrates robustness and introduces fewer noisy labels during implementation, thereby enhancing the quality and effectiveness of the labeling process.

\begin{figure*}[t!]
\centering
\includegraphics[width=\linewidth]{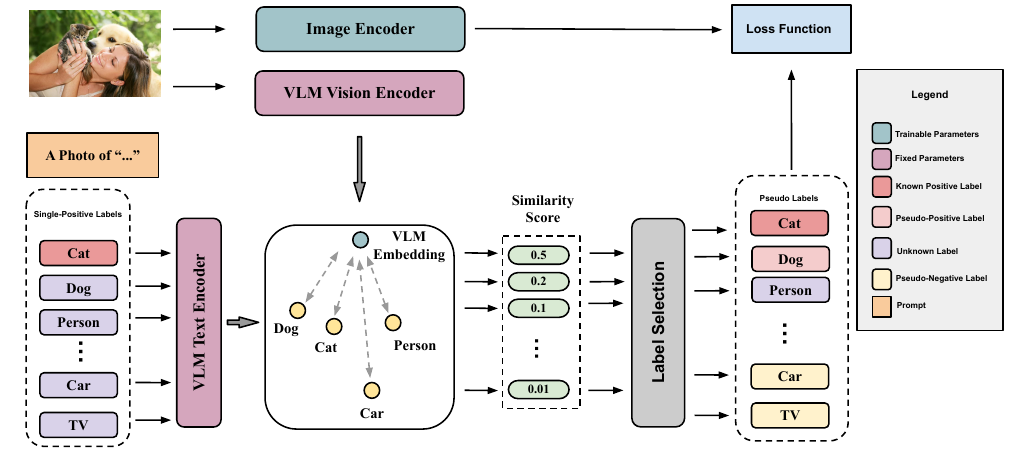}
\caption{The architecture of the proposed model. The image encoder is a trainable model for feature extraction, such as ResNet and Vision Transformer, meanwhile, the VLM Vision and Text encoders are fixed for vision-language embedding space. By the vision-language embedding operation, instead of the original single-positive label offering, we can access a brand new pseudo-label as the reference for the final prediction.}
\label{fig_architecture}
\end{figure*}

\subsection{Vision-Language Model}
Vision-Language Model (VLM)~\cite{radford2021learning, jia2021scaling} is a multi-modality model, using the image and its corresponding text as supervision signal to help us better understand vision-language correlation. The most commonly used VLM model is named Contrastive Language-Image Pre-Training (CLIP)~\cite{radford2021learning}. Since CLIP is a well-trained vision-language model, it has impressive potential as the tool for different downstream tasks including multi-label learning. In DualCoOp~\cite{sun2022dualcoop}, the authors leverage CLIP for multi-label recognition tasks in limited-annotation domains (although not single-label). They highlight the benefit of learning the relationship between different category names in the multi-label recognition task and observe that the aligned image and textual CLIP spaces can be used for this purpose. Concretely, they learn a positive and negative ``prompt context'' -- a sequence of embedding vectors -- for possible target category names. These prompt contexts can then be used as classifiers by computing the similarity between local features in an image and each of the context vectors, and assigning a positive or negative label for each category (at each location) based on whichever context has the higher similarity. The current state-of-the-art SPML approach is SCPNet~\cite{ding2023exploring}, which proposes to explore structured semantic prior information to better understand label-to-label associations in images. The authors use a CLIP model to extract an object association matrix as the prior information to achieve better performance. MKT~\cite{he2023open} is proposed for the zero-shot multi-label learning by applying the knowledge transfer model with CLIP model's initial weights. In our study, we propose a simpler approach to leveraging CLIP features, called Vision-Language Pseudo-Label (VLPL), that works in the single-label annotation domain and only requires the selection of positive pseudo-labels.


\section{Approach}
 
In this section, we describe the problem definition, detail the architecture of the proposed model, illustrate the vision language pseudo-labeling method, and describe our loss function. 

\subsection{Problem Definition}
In the context of multi-label learning (MLL), we are given a dataset $D=\{X_{i}, Y_{i}\}^{N}_{i=1}$ consisting of $N$ training samples. Each sample $X_{i}$ is an input image, and its corresponding label vector $Y_{i} \in \{-1, 1\}^{L}$ has a length $L$. Within this label vector, $Y_{il}=1$ designates a positive label that is relevant to $X_{i}$, whereas $Y_{il}=-1$ signifies a negative label that is irrelevant to $X_{i}$. Our study focuses on Single-Positive Multi-Label learning (SPML), where there exists only one positive label and all others are unknown. To indicate this, we modify the label vector annotation as $Y_{i}\in \{-1, \emptyset, 1\}^{L}$. The symbol $\emptyset$ in $Y_{il}$ indicates that the association of the $l-th$ label with the input image $X_{i}$ remains undetermined. 

We have the annotation $\Sigma_{l=1}^L\mathds{1}_{y_{il}}=1$, where $\mathds{1}_{[\cdot]}$ represents an indicator function.  This implies that each input image has only one observed positive label, and the rest remain unknown. The main objective of the SPML study is to learn a mapping function $f: X \rightarrow Y$ from the dataset $D$.  The ground truth label is $Y=\{-1, 1\}$, while the observed label is limited to $Y^{'}=\{\emptyset, 1\}$, making SPML a challenging task within the realm of MLL, operating under limited supervision.

\subsection{Architecture}
In Figure~\ref{fig_architecture}, we present the architecture of our proposed model. This model is designed with two branches: the upper branch is comprised of a trainable feature extraction component, with the flexibility to use any image encoder as its backbone. For further details about the specific image encoder utilized in our experiments, please refer to Section~\ref{sec:evaluation}. The lower branch includes the fixed VLM, which we use for pseudo-label prediction. We initialize a Vision Encoder and Text Encoder with pre-trained CLIP weights, keeping these weights fixed throughout the process. These encoders produce 768-dimensional output embeddings. To generate embeddings for each possible image label, we compute the CLIP text embedding based on the prompt ``A photo of $X$''. As the text encoder remains static, these embeddings only have to be computed once. During the model's operation, the input image is fed into both the trainable image encoder and the fixed CLIP vision encoder, resulting in visual embeddings. This CLIP visual embedding can be compared with the text embeddings generated from text prompts in the form of ``A Photo of ...". To determine pseudo-label assignments, we compute the cosine similarity between the visual and label embeddings from the text prompts. Based on this similarity measure, we select pseudo-labels for each image. Subsequently, we utilize these pseudo-labels, in conjunction with the single positive label, as the final reference for training the entire model, enabling it to make accurate predictions in tasks.

\subsection{Vision Language Pseudo-Labeling}

\begin{algorithm}
\caption{Vision Language Pseudo-Labeling}
\label{alg:algorithm}
\textbf{Input}: The visual embedding $e_{I}$ and the query label embedding $e_{L}^{i}$\\
\textbf{Parameter}: positive pseudo-label threshold $\theta$, negative pseudo-label partial $\delta \%$ \\
\textbf{Output}: Pseudo-Labeling $Y$ of input image 
\begin{algorithmic}[1] 
\STATE Compute the image and the query labels similarity as equation 1 
\STATE Let $i=0$.
\WHILE{$i<L-1$}
\IF {$p_{i}> \theta$}
\STATE $Y_{i}=1 (Positive-label)$
\ELSE
\STATE $Y_{i}= \emptyset (Unknow-label)$
\ENDIF
\STATE $i=i+1$
\ENDWHILE
\STATE Rank the similarity vector $P=[p_{1},...,p_{L}]$ and set \\ $\delta \%$ of the smallest similarity values as negative labels.
\STATE \textbf{return} $Y$ 
\end{algorithmic}
\end{algorithm}

Our pseudo-labeling method benefits from the ``open-world" capabilities of the large VLM, enabling the use of rich, free-form text with a long list vocabulary of visual categories. In zero-shot learning applications, these VLMs allow us to obtain visual embeddings and potential label embeddings with ease. The cosine similarity between these embeddings can then be used to perform image classification. We adopt this methodology for multi-label learning in our work. As shown in Figure~\ref{fig_architecture}, the VLPL employs a visual encoder $E_{V}: R^{w \times h \times 3} \rightarrow R^{d}$ and a text encoder $E_{L}: R^{m \times d_{c}} \rightarrow R^{d}$ to extract the image and text embeddings, respectively. The visual inputs are 3-channel images of shape $w \times h$, while the text inputs are prompts consisting of $m$ words, each of which is embedded into a $d_{c}$-dimensional vector. Both the image and text inputs are mapped into a $d$-dimensional latent space. We obtain a visual embedding vector $e_{I} \in R^{d}$ and $n$ label embedding vector $e_{L}^{1}, e_{L}^{2}, ..., e_{L}^{n} \in R^{d}$, where $n$ denotes the label number of the category space. We compute the dot product between $e_{I}$ and each of the $e_{L}^{i}$, resulting in an $n$-dimensional vector, where the $i$-th element means the similarity between the image and the $i$-th label query. This similarity vector can be used to predict the label of the image, and the probability $p_{i}$ of the appearance of the $i$-th label on the image is computed by the temperature softmax function
\begin{equation}
\label{equ:1}
p_{i}=\frac{exp(<e_{I},e_{L}^{i}>/ \tau)}{\Sigma_{j=1}^{L}exp(<e_{I},e_{L}^{j}>/ \tau)}
\end{equation}
where $<\cdot, \cdot>$ denotes the dot product and $\tau$ is a temperature scalar of the softmax function.

As shown in Algorithm~\ref{alg:algorithm}, we use the equation~\ref{equ:1} to measure the similarity between the image and query labels. We determine the pseudo-label in three formations: positive, negative, and unknown using one threshold, namely, a positive threshold $\theta$ and a negative label percentage coefficient $\delta \%$. For the input image $I$ and the $i$-th query label, if the measurement similarity $p_{i}$ is over than $\theta$, we set $Y_{i}$ as positive. In terms of the large label space of multi-label learning, a given image has a few positive labels, and the rest are negative. We can rank the similarity vector $P=[p_{1},...,p_{L}]$ and set $\delta \%$ of the smallest similarity values as negative labels. Afterward, the rest are set unknown. By following this simple algorithm, we can generate a new pseudo-label vector given the input image.

\subsection{Loss Function}
In the SPML domain, there is only one ground truth positive label, and the rest are unknown. How these labels are utilized by the loss function plays a crucial role in the model training. The Assuming-Negative (AN) Loss, where the unknown labels are assumed to be negative, is commonly used as the baseline loss for SPML tasks, but leads to the generation of many false negatives during the model training. We instead use the more recently proposed Entropy-Maximization (EM) loss, which acknowledges the un-annotated labels as unknown, rather than negative, and seeks to maximize the entropy of predicted probabilities for the unknown labels.

\begin{equation}
\label{equ:3}
\begin{split}
Loss_{EM}(x^{(n)}, y^{(n)})= -\frac{1}{L}\Sigma_{l=1}^{L} 
[\mathds{1}_{[y_{l}^{n}=1]}log(f_l(x^{(n)})) \\
+ \mathds{1}_{[y_{l}^{n}=\emptyset]}\alpha H(f_l(x^{(n)}))]
\end{split}
\end{equation}

\begin{equation}
\label{equ:4}
\begin{split}
H(f_l(x^{(n)})) =-[f_l(x^{(n)})log(f_l(x^{(n)})) \\ 
+(1-f_l(x^{(n)}))log(1-f_l(x^{(n)}))]
\end{split}
\end{equation}
where $H(f(x^{(n)}))$ is the entropy loss of the unknown labels.

In our model, the VLPL will generate pseudo-labels for positive, negative, and unknown categories. We use the EM loss strategy to acknowledge the unknown labels and integrate the pseudo-label loss of our model. Our loss is the following:

\begin{equation}
\label{equ:5}
\begin{split}
Loss(x^{(n)}, y^{(n)})=-\frac{1}{L}\Sigma_{l=1}^{L}[\mathds{1}_{[y_{l}^{n}=1]}log(f_l(x^{(n)})) \\ 
+\mathds{1}_{[y_{l}^{n}=\emptyset]}\alpha H(f_l(x^{(n)}))\\ 
+ \mathds{1}_{[y_{l}^{n}=\hat{1}]}\beta S(f_l(x^{(n)}))   
+ \mathds{1}_{[y_{l}^{n}=-\hat{1}]}\gamma S(f_l(x^{(n)}))]
\end{split}
\end{equation}

\begin{equation}
\label{equ:6}
S(f(x^{(n)})) = (1-\rho)log(1-f_l(x^{(n)}))-\rho log(f_l(x^{(n)}))
\end{equation}
where, $\mathds{1}_{[y_{l}^{n}=\hat{1}]}$ denotes the pseudo positive-label, $\mathds{1}_{[y_{l}^{n}=-\hat{1}]}$ denotes the pseudo negative-label, $S(f(x^{(n)}))$ is the  pseudo-label loss with labeling smooth $\rho$, $\alpha$, $\beta$, $\gamma$ are the coefficient of each loss section. 

However, in experimentation (discussed in Section~\ref{sec:pnl}), we found that the best model performance is achieved when we only use pseudo-positive labels, while keep the rest labels unknown, rather than including pseudo-negative labels. Therefore our final loss function is:
\begin{equation}
\label{equ:7}
\begin{split}
Loss(x^{(n)}, y^{(n)})=-\frac{1}{L}\Sigma_{l=1}^{L}[\mathds{1}_{[y_{l}^{n}=1]}log(f_l(x^{(n)})) \\ 
+\mathds{1}_{[y_{l}^{n}=\emptyset]}\alpha H(f_l(x^{(n)}))
+ \mathds{1}_{[y_{l}^{n}=\hat{1}]}\beta S(f_l(x^{(n)}))   
\end{split}
\end{equation}
\section{Evaluation}
\label{sec:evaluation}
\begin{table*}[h!]
\center
\begin{tabular}{l|c|cccc}
\hline\noalign{\smallskip}
Ann. Labels &Methods                &VOC &COCO &NUS &CUB   \\
\noalign{\smallskip}
\hline
\noalign{\smallskip}
\multicolumn{6}{c}{\textit{Full Annotation}}\\
All P. \& All N &BCE Loss     & $89.42$ &$76.78$ &$52.08$ & $30.90$  \\
\hline\hline
\multicolumn{6}{c}{\textit{Limited Annotation}}\\
1 P. \& All N   &BCE Loss       &$87.60$ &$71.39$ &$46.45$ & $20.65$  \\
\hline 

 &AN Loss      &$85.89$ &$64.92$ &$42.27$ & $18.31$  \\
 &DW~\cite{cole2021multi}                &$86.98$ &$67.59$ &$45.71$ & $19.15$  \\
 &L1R~\cite{cole2021multi}             &$85.97$ &$64.44$ &$42.15$ & $17.59$   \\
 &L2R ~\cite{cole2021multi}            &$85.96$ &$64.41$ &$42.72$ & $17.71$   \\
 &LS~\cite{muller2019does}                 &$87.90$ &$67.15$ &$43.77$ & $16.26$   \\
 &N-LS~\cite{cole2021multi}           &$88.12$ &$67.15$ &$43.86$ & $16.82$   \\
1 P. \& 0 N. &EntMin~\cite{grandvalet2004semi}      &$53.16$ &$32.52$ &$19.38$ &$13.08$   \\
 &Focal Loss~\cite{lin2017focal}   &$87.59$ &$69.79$ &$47.00$ &$19.80$   \\
 &ASL~\cite{ridnik2021asymmetric}      &$87.76$ &$68.78$ &$46.93$ &$18.81$   \\
 &ROLE~\cite{cole2021multi}     &$87.77$ &$67.04$ &$41.63$ &$13.66$   \\
 &ROLE+LI~\cite{cole2021multi}    &$88.26$ &$69.12$ &$45.98$ &$14.86$   \\
 &EM~\cite{zhou2022acknowledging}        &$89.09$ &$70.70$ &$47.15$ & $20.85$  \\
 &EM+APL~\cite{zhou2022acknowledging}    &$89.19$ &$70.87$&$47.59 $ & $21.84$  \\
 &LL-R~\cite{kim2022large}      &$\mathbf{89.2}$ &$71.0$&$47.4 $ & $19.5$  \\
 &LL-Ct~\cite{kim2022large}     &$89.0$ &$70.5$&$48.0 $ & $20.4$  \\
 &LL-Cp~\cite{kim2022large}     &$88.4$ &$70.7$&$48.3 $ & $20.1$  \\
 \hline
1 P. \& 0 N. & VLPL(Ours) &$89.10$ &$\mathbf{71.45}$ &$\mathbf{49.55}$ &$\mathbf{24.02}$ \\ 
\hline
\end{tabular}
\caption{Results of the different models with the same experimental settings as the~\cite{cole2021multi}. Using the same input image size setting $448 \times 448$, our model outperforms the limited-annotation baseline models for the benchmark COCO, NUS-WIDE, and CUB.}
\center
\label{table_idea}
\end{table*}

\subsection{Implementation Details}
Our models are implemented using PyTorch. We train the model for 10 epochs, using the Adam~\cite{kingma2014adam} optimizer. The batch size is 8. The learning rate is determined by grid search in the range of the \{ $1e-3$,$1e-4$,$1e-5$\}, and we find $1e-5$ yields the best performance. For data augmentation, we use horizontal flipping for the training dataset with 50\% probability. Images are resized to $448 \times 448$ for both our proposed model and all baseline models. We follow the conventions of previous works in multi-label classification~\cite{cole2021multi,zhou2022acknowledging} for model evaluation and report the mean average precision (mAP).

\subsection{Dataset}

Since there are currently no datasets explicitly designed for Single Positive Multi-label Learning (SPML), we use~\cite{cole2021multi} and others' adaptation of existing large-scale multi-label datasets to simulate a ``single positive" scenario. This method allows us to retain all ground-truth labels for performance evaluation and training phenomena analysis. After setting aside 20\% of the training images for validation, one random positive label is kept for each training image, treating all other labels as un-annotated. This operation is performed once for each dataset. It's important to note that the validation and test sets remain fully labeled. We use four well-known datasets in our experiments, namely: PASCAL VOC (VOC)~\cite{pascal-voc-2012}, MS-COCO (COCO)~\cite{lin2014microsoft}, NUS-WIDE(NUS)~\cite{chua2009nus}, and CUB-200-2011(CUB)~\cite{wah2011caltech}.

\noindent
\textbf{PASCAL Visual Object Classes Challenge (VOC2007)}~\cite{pascal-voc-2012} is a widely used dataset for multi-label recognition. It contains 5,011 images in the training/validation set, and 4,952 images as the test set. There are 20 possible classes, with an average of 2.5 categories per image.

\noindent
\textbf{Microsoft COCO}~\cite{lin2014microsoft} (MS-COCO) is another widely used benchmark for multi-label image recognition. It contains 82,801 training images and 40,504 validation images. There are 80 categorized objects in this dataset, with an average of 2.9 object labels per image. Since this data set lacks of test set, the validation images are often used for evaluation in the literature.

\noindent
\textbf{NUS-WIDE}~\cite{chua2009nus} is a real-world web image dataset. Originally, the dataset contains 269,648 Flickr images with 81 manually annotated visual concepts. However, due to some Flickr image downloading links expiring, it is impractical to evaluate our model using the original dataset. We use the curated dataset provided by the authors of AckUnknown~\cite{zhou2022acknowledging}, which has 150,000 training images and 60,260 test images, to conduct our experiment and provide a fair comparison to recent work. 

\noindent
\textbf{CUB}~\cite{wah2011caltech} is a dataset for fine-grained visual categorization task. It contains 11,788 images of 200 subcategories belonging to birds, with 5,994 training images and and 5,794 testing images. In the experiment, the model will predict 312 attributes of each bird images. 

\begin{table*}[h!]
\center
\begin{tabular}{l|l|cccc}
\hline
Backbone  & Pretrained &VOC &COCO &NUS &CUB\\
\hline 
\multicolumn{2}{c|}{Current SOTA (EM+APL~\cite{zhou2022acknowledging})} &$89.19$ &$70.87$&$47.59 $ &$21.84$    \\
\hline
ResNet-50 (Ours)  &ImageNet1k &$89.10$ &$71.45$ &$49.55$ &$24.02$ \\
ConvNeXt-XL (Ours)    &ImageNet1k &$93.31$ &$83.37$  &$56.11$  &$25.49$ \\
ConvNeXt-XL (Ours)  &ImageNet22k &$93.37 (4.6\% \uparrow)$ &$\mathbf{84.65}(18.4\% \uparrow)$ & $\mathbf{57.12}(15.2\% \uparrow)$ &$\mathbf{26.04}(8.4\% \uparrow)$\\
ViT-L (Ours)     & CLIP &$\mathbf{94.16} (5.5\% \uparrow)$ &$80.55$ & $52.53$ & $12.82$ \\
\hline
\end{tabular}
\caption{Results of the different network architectures and pretraining initializations for the proposed model. Compared with ResNet50 backbone, all of the chosen large network architectures show better performance. Further, it shows the ConvNeXt-XL with ImageNet22k pretraining initialization weights achieves comparable performance across the different datasets.} 
\center
\label{table_network}
\end{table*}

\subsection{Baseline Performance}
\label{sec:baseline_performance}
Most of the current state-of-the-art models use a ResNet50 backbone. Because of this, we first explore the performance of our proposed model using a ResNet50. Our experimental setting -- using a single-positive label adaptation of the datasets and reporting the mAP evaluation metric, evaluated on the model that achieves the highest accuracy on a withheld validation set -- is the same as the previous methods~\cite{cole2021multi, zhou2022acknowledging, kim2022large}. Table~\ref{table_idea} reports the performance results of the different models.  The proposed model, VLPL, achieves $mAP=89.10$ on the VOC dataset, $mAP=71.45$ on the COCO dataset, $mAP=49.55$ on the NUS-WIDE dataset, and $mAP=24.02$ on the CUB dataset. Compared with the SOTA baselines under the same experimental setting, VLPL demonstrated superior performance across all the benchmarks, indicating the effectiveness of our proposed model. While we focus on the limited annotation setting, we also compare our performance to a model trained using full ground truth annotations. Our performance using limited annotations is competitive with the model using the full set of ground truth annotations.

Note that we do not compare to SCPNet~\cite{ding2023exploring}. The SCPNet paper reports on an exponentially weighted moving average of all possible models. We do not believe this is a fair comparison, given that all of the other papers in this space select a single model to evaluate based on the performance of the withheld validation set. Additionally, there is currently no publicly available code for SCPNet, preventing us from evaluating their approach in the same experimental setting as all of the other compared approaches. 

\subsection{Backbone Experimentation}
In the previous section, we prove the effectiveness of our proposed model using a ResNet50 model. The current state-of-the-art methods for SPML concentrate more on the methodology but ignore the power of the backbone architecture. It's widely known, however, that for computer vision tasks, an optimal network architecture with suitable pretraining initialization can substantially improve the model's performance~\cite{ridnik2021imagenet}. To this end, we explored various network architectures and pretraining initializations. In our implementation, we select ConvNeXt-XL and ViT-L as the backbone. For the pretraining initialization, we choose the ConvNeXt-XL model pretrained under ImageNet1k and ImageNet22k, and the ViT-L model initialized with CLIP weights. Table~\ref{table_network} presents the different pretrained backbone performances. We observe that all the chosen large network architectures outperform the ResNet50 backbone, indicating that model architecture scaling can be beneficial for performance improvement. While the model scale has a significant impact on performance, the performance gap between different pretraining initializations is not significant.

While the ResNet50 baseline trained using VLPL already outperformed other SOTA methods, the best larger models achieve even more impressive performance improvement. Specifically, we achieved a $5.5\%$ increase in mAP (up to $94.16$) on the VOC dataset, a $18.4\%$ increase (up to $84.65$) on the COCO dataset, a $15.2\%$ increase (up to $57.12$) on the NUS-WIDE dataset, and an $8.4\%$ increase (up to $26.04$) on the CUB dataset.  

Considering the model performance across all four benchmarks, the ConvNeXt-XL model pretrained with ImageNet-22k achieves the best performance on average. Though the large backbones show promising performance improvement, we use ResNet50 as the backbone in the following experiments to ensure a fair comparison with previous works.  

\begin{figure*}[t!]
\centering
\includegraphics[width=\linewidth]{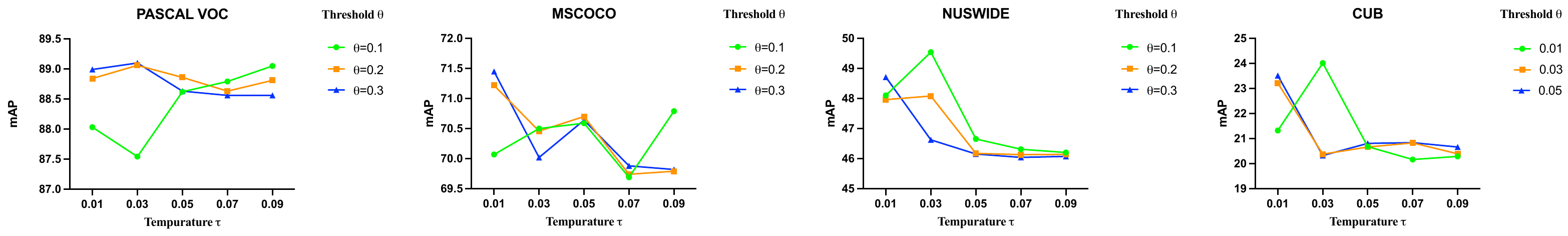}
\caption{The hyperparameter search of the temperature scalar $\tau$, and pseudo-labeling threshold $\theta$ across the four benchmark datasets: PASCAL VOC, COCO, NUS-WIDE, and CUB.}
\label{fig_hyperparameter}
\end{figure*}

\subsection{Positive Labels vs. Negative Labels}
\label{sec:pnl}
In this section, we explore the influence of pseudo-positive and pseudo-negative labels on the performance of the VLPL model. Multi-label learning datasets are typically characterized by an imbalance between positive and negative labels, as noted in previous studies~\cite{Ridnik_2021_ICCV}. For instance, the Pascal VOC dataset has 20 potential labels, yet the average image contains only 2.5 positive labels. This imbalance creates a challenging environment for model learning. In the context of our method, for a predicted label vector, we can predict pseudo-negative labels by ranking the similarity scores between visual and label embeddings. We then select the labels with the least $k\%$ similarity scores as pseudo-negative labels. The rationale behind this is that the labels with the least similarity scores are likely to be the ones that are most irrelevant or `negative' to the given image. To understand the impact of the number of pseudo-negative labels on model performance, we conducted experiments using varying percentages of pseudo-negative labels, ranging from 10\% to 40\% of the total label count on the benchmark Pascal VOC dataset.

\begin{table}

\centering

\begin{tabular}{l|cccc}
\hline
$\delta \%$ &$10\%$  &$20\%$ &$30\%$  &$40\%$ \\
\hline
Pos. + Neg. & $88.51$  & $88.43$  & $88.33$   & $88.13$     \\
\hline
Pos. only & \multicolumn{4}{c}{$89.10$}\\
\hline
\end{tabular}

\caption{The results on different pseudo-negative label percentages over the total label vector (from $10\%$ to $40\%$). We conducted the experiments on benchmark PASCAL VOC.}
\label{table_negativelabel}
\end{table}

Table~\ref{table_negativelabel} shows the results of model performance on different pseudo-negative labels. We observe a decrease in model performance as the number of pseudo-negative labels increases. Our hypothesis for the cause of this is that the imbalance between positive and negative labels negatively impacts model performance. By comparing models that use both pseudo-positive and pseudo-negative labels against models that only utilize pseudo-positive labels it appears that the latter approach provides better performance. In light of these findings, our practical implementation uses pseudo-positive labeling exclusively, while treating the remaining labels as unknown.

\subsection{Label Smoothing}
\begin{table}[h!]
\centering
\begin{tabular}{l|cccc}
\hline
          &VOC & COCO &NUS &CUB  \\
\hline
w/o LS    &  88.69 & 70.96  & 49.13 & 23.71 \\
w LS      &  89.10 & 71.34  & 49.55 & 24.02 \\
\hline
\end{tabular}

\caption{The results on model performance with and without label smooth (LS) on the pseudo-labeling. We conducted the experiments on four benchmarks. The results indicate that LS contributes more to the model performance.}
\label{table_labelsmooth}
\end{table}    
Label smoothing (LS) is a method to overcome overfitting and mitigate the label noise for multi-class classifiers~\cite{muller2019does, szegedy2016rethinking}. In our implementation, we adopt LS for our pseudo-labeling and set the $\epsilon=0.9$. Therefore, the $i$-th pseudo positive-label category entropy loss function is $loss_{i}=-[\epsilon (log(f(x))) + (1-\epsilon)(log(1-f(x))]$. As shown in Table~\ref{table_labelsmooth}, we conduct the experiments with and without LS over four benchmarks. The results indicate applying LS improves the model performance compared to the model without it. We set the label smoothing as the default setting in our whole experiment.

\begin{figure*}[t!]
\centering
\includegraphics[width=\linewidth]{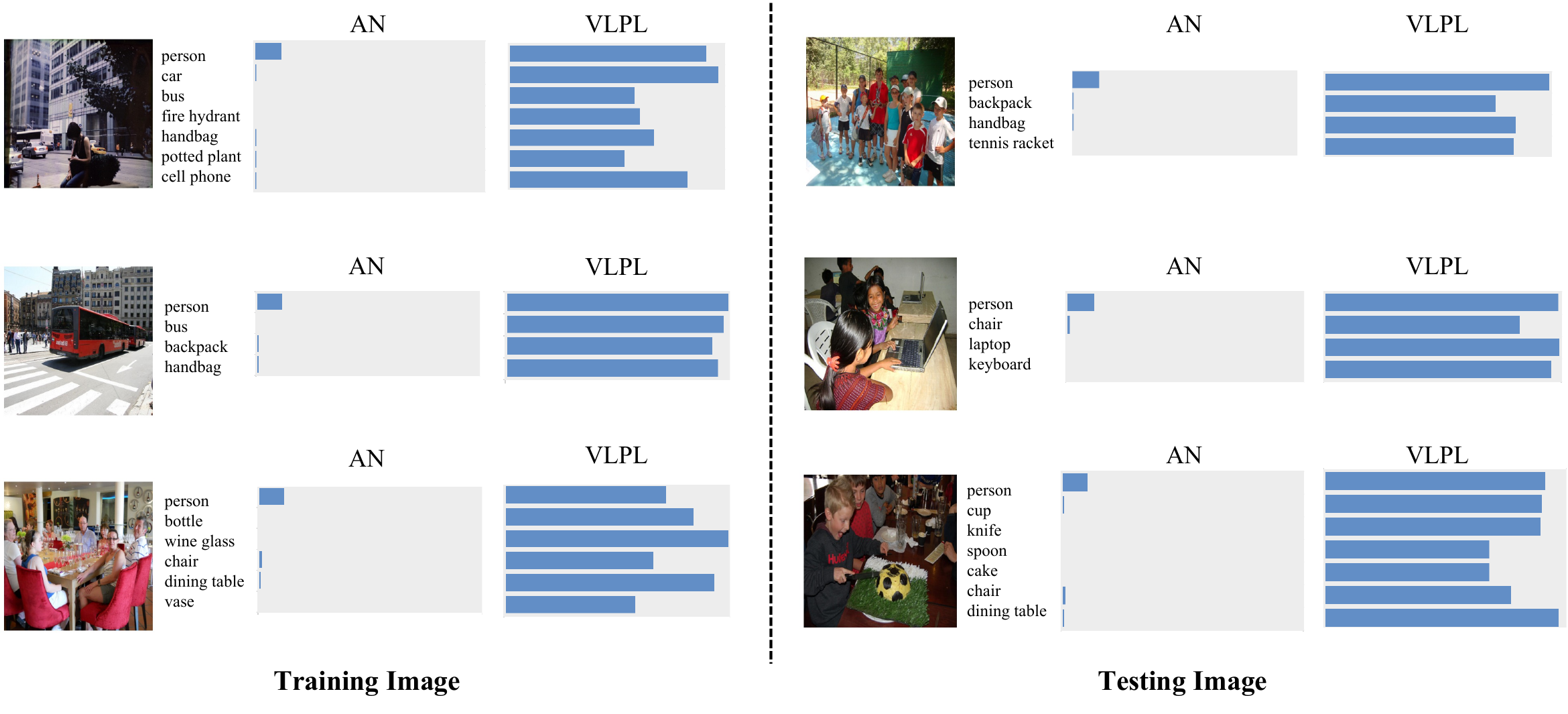}
\caption{The visualization results on the training and testing images of COCO dataset. The blue bar is the prediction probability of each positive label. Compared with the baseline method AN, our VLPL shows superior performance for the final label prediction.}
\label{visulization_example}
\end{figure*}

\subsection{Temperature Hyperparameter \& Threshold}
The temperature scalar $\tau$ of equation~\ref{equ:1} and pseudo-labeling threshold $\theta$ are crucial hyperparameters, as they directly influence the pseudo-labeling processing. We conducted a hyperparameter sweep
for these two hyperparameters. Considering the different benchmark datasets under different label query numbers, especially for the CUB dataset. We set $\tau = [0.01, 0.03, 0.05, 0.07, 0.09]$ and $\theta = [0.1, 0.2, 0.3]$ for VOC, COCO, and NUS-WIDE datasets.
Meanwhile, we set $\tau = [0.01, 0.03, 0.05, 0.07, 0.09]$ and $\theta = [0.01, 0.03, 0.05]$ for CUB to identify the optimal settings. Our results, shown in Figure~\ref{fig_hyperparameter}, indicate that the model performance is sensitive to the $\tau$ and $\theta$. 
To visualize the performance of the different hyperparameters, we set each threshold $\theta$ as the condition and plot the different temperature value $\tau$ over the same $\theta$.
In terms of the performance over different benchmarks, $\tau=0.03$ and $\theta=0.3$ yield superior performance for VOC ($mAP=89.10$) dataset, $\tau=0.01$ and $\theta=0.3$ yield superior performance for COCO ($mAP=71.45$) dataset, $\tau=0.03$ and $\theta=0.1$ yield superior performance for NUS-WIDE ($mAP=49.55$), and $\tau=0.03$ and $\theta=0.01$ yield superior performance for CUB ($mAP=24.02$).

\subsection{Results Visualization}
To demonstrate the efficacy of our approach, we provide visualizations in Figure~\ref{visulization_example}, highlighting the label prediction probability generated by our method compared with those from the baseline model (referred to as AN). Our proposed model demonstrates a high level of confidence in accurately identifying positive labels, compared with the predictions offered by the baseline model. For the purposes of our visualization, we have intentionally selected instances from both the training and testing datasets. This choice allows us to showcase how our VLPL method consistently excels in performance across different stages of model training and testing. The visual evidence serves as a testament to the advantages of employing VLPL, both during the training phase and in the application to unseen data.

\section{Conclusion}
In this study, we introduced an innovative yet approach to single positive multi-label learning, named VLPL. VLPL leverages a large-scale vision-language model and utilizes the aligned visual and textual embedding similarities to generate pseudo-labels. Our method consists of simple components and results in significant performance improvements across several popular datasets when compared to existing, more complex approaches. To better understand the impact of various factors within the VLPL framework and explore how to maximize accuracy, we carried out a comprehensive set of experiments and ablations. 



{\small
\bibliographystyle{ieee_fullname}
\bibliography{main}
}

\end{document}